\documentclass[lettersize,journal]{IEEEtran}
\usepackage{multirow}
\usepackage{amsmath,amsfonts}
\usepackage{algorithmic}
\usepackage{algorithm}
\usepackage{array}
\usepackage[caption=false,font=normalsize,labelfont=sf,textfont=sf]{subfig}
\usepackage{textcomp}
\usepackage{stfloats}
\usepackage{url}
\usepackage{verbatim}
\usepackage{graphicx}
\usepackage{cite}
\usepackage{color,xcolor}
\usepackage{amsthm}
\newtheorem{remark}{Remark}
\begin{document}

\title{Adaptive Pruning for Large Language Models with Structural Importance Awareness}
\author{Haotian Zheng, Jinke Ren, Yushan Sun, Ruichen Zhang, Wenbo Zhang, Zhen Li, \\Dusit Niyato,~\IEEEmembership{Fellow,~IEEE,}
Shuguang Cui,~\IEEEmembership{Fellow,~IEEE,}
and Yatong Han
\thanks{
The work was supported in part by NSFC with Grant No. 62293482, the Basic Research Project No. HZQB-KCZYZ-2021067 of Hetao Shenzhen-HK S\&T Cooperation Zone, the Shenzhen Outstanding Talents Training Fund 202002, the Guangdong Research Projects No. 2017ZT07X152 and No. 2019CX01X104, the Guangdong Provincial Key Laboratory of Future Networks of Intelligence (Grant No. 2022B1212010001), and the Shenzhen Key Laboratory of Big Data and Artificial Intelligence (Grant No. ZDSYS201707251409055). \it{Haotian Zheng and Jinke Ren contributed equally to this work.}
\it{(Corresponding authors: Jinke Ren and Yatong Han.)}}
\thanks{H. Zheng is with the National Key Laboratory of Autonomous Marine Vehicle Technology, Harbin Engineering University, Harbin 150001, China, and also with the Shenzhen Future Network of Intelligence Institute (FNii-Shenzhen), The Chinese University of Hong Kong, Shenzhen 518172, China (e-mail: 13703689922@hrbeu.edu.cn).}
\thanks{J. Ren is with the FNii-Shenzhen, the School of Science and Engineering (SSE), and the Guangdong Provincial Key Laboratory of Future Networks of Intelligence, The Chinese University of Hong Kong, Shenzhen 518172, China (e-mail: jinkeren@cuhk.edu.cn).}
\thanks{Y. Sun is with the National Key Laboratory of Autonomous Marine Vehicle Technology, Harbin Engineering University, Harbin 150001, China (e-mail: sunyushan@hrbeu.edu.cn).}
\thanks{R. Zhang and D. Niyato are with the College of Computing and Data Science, Nanyang Technological University, Singapore (e-mail:
ruichen.zhang@ntu.edu.sg; dniyato@ntu.edu.sg).}
\thanks{W. Zhang is with the Aerospace Science and Industry Shenzhen (Group) Co., Ltd, Shenzhen 518048, China (e-mail: 12032717@mail.sustech.edu.cn).}
\thanks{Z. Li and S. Cui are with the SSE, the FNii-Shenzhen, and the Guangdong Provincial Key Laboratory of Future Networks of Intelligence, The Chinese University of Hong Kong, Shenzhen 518172, China (e-mail: lizhen@cuhk.edu.cn; shuguangcui@cuhk.edu.cn).}
\thanks{Y. Han is with the FNii-Shenzhen and the Guangdong Provincial Key Laboratory of Future Networks of Intelligence, The Chinese University of Hong Kong, Shenzhen 518172, China, and also with Infused Synapse AI, Shenzhen 518048, China (e-mail: hanyatong@cuhk.edu.cn).}}
\maketitle

\begin{abstract}
The recent advancements in large language models (LLMs) have significantly improved language understanding and generation capabilities. However, it is difficult to deploy LLMs on resource-constrained edge devices due to their high computational and storage resource demands. To address this issue, we propose a novel LLM model pruning method, namely structurally-aware adaptive pruning (SAAP), to significantly reduce the computational and memory costs while maintaining model performance. We first define an adaptive importance fusion metric to evaluate the importance of all coupled structures in LLMs by considering their homoscedastic uncertainty. Then, we rank the importance of all modules to determine the specific layers that should be pruned to meet particular performance requirements. Furthermore, we develop a new group fine-tuning strategy to improve the inference efficiency of LLMs. Finally, we evaluate the proposed SAAP method on multiple LLMs across two common tasks, i.e.,  zero-shot classification and text generation. Experimental results show that our SAAP method outperforms several state-of-the-art baseline methods, achieving 2.17\%, 2.37\%, and 2.39\% accuracy gains on LLaMA-7B, Vicuna-7B, and LLaMA-13B. Additionally, SAAP improves the token generation speed by 5\%, showcasing its practical advantages in resource-constrained scenarios.
\end{abstract}

\section{Introduction}
\IEEEPARstart{I}n the past two years, large language models (LLMs) have become the leading solution for many practical applications, such as finance, medicine, and education, due to their powerful natural language understanding and generation capabilities \cite{ref1}. However, the massive size of LLMs, often consisting of hundreds of billions to trillions of parameters, results in high computational latency and low memory efficiency \cite{ref2, ref3}. This makes real-time processing and flexible scalability challenging, especially for the practical deployment of LLMs on resource-constrained edge devices \cite{zhang_1,ref4}. To address this issue, lightweight deployment of LLMs has become a key research direction to enhance LLMs' accessibility across diverse platforms \cite{ref5}.

Recently, model pruning has been recognized as a promising solution to reduce LLMs' model size and computational overhead while maintaining their model performance \cite{ref7}. Specifically, model pruning reduces computational complexity by removing unnecessary weights or structures from a model without sacrificing the model’s key functionality and prediction accuracy \cite{ref8}. Moreover, by focusing on important structures, model pruning can also mitigate overfitting issues often present in large models, particularly LLMs \cite{ref9}. Thus far, many pioneering studies have emphasized the importance of structured pruning to balance model performance and resource efficiency \cite{zhang_2,ref11}. In particular, several advanced pruning techniques have been developed to adaptively remove weights based on their contributions to model performance \cite{ref12,ref13}. 

Despite these advancements, there remain three challenges in LLM pruning: 1) {\it Weight importance estimation}, where accurately estimating weight importance is crucial to pruning without affecting model performance; 2) {\it Layerwise pruning ratio}, where a uniform ratio may not be suitable for all structures in LLM; and 3) {\it Fine-tuning}, where fine-tuning pruned LLMs is essential for recovering their performance \cite{zhang_3}. Several early studies have explored pruning methods that rely on uniform metrics, typically using single or linear approaches \cite{ref15,ref16,ref17,llm-pruner}. However, these metrics often oversimplify pruning decisions and fail to capture the intricate interdependencies of coupled structures. On the other hand, post-pruning fine-tuning is crucial to restore accuracy but consumes significant computational and storage resources. Therefore, achieving an optimal balance between memory efficiency and model performance remains a challenge in LLM pruning.

To address this issue, we introduce {\it structurally-aware adaptive pruning (SAAP)}, a novel method designed to improve LLM pruning by selectively removing non-essential structures while reducing computational and memory usage. SAAP employs an adaptive metric to assess structural importance and prunes these structures that exhibit instability under varying conditions. Furthermore, it employs group-wise fine-tuning in the recovery stage to maintain model performance. Instead of relying solely on importance scores, SAAP considers fluctuations, providing a precise and efficient approach for structured pruning. Experimental results demonstrate the superiority of SAAP over several baseline methods. The main contributions of this paper are summarized as follows.
\begin{itemize}
    \item  We propose an adaptive importance fusion metric to accurately estimate weight importance. By adopting the importance scores of different structures, SAAP can be optimized at various layers and stages in different LLMs.
    \item We introduce an adaptive structure search approach to achieve layerwise pruning. By calculating the stability of the importance score of each coupled structure, we provide a unified evaluation system for assessing the importance of model parameters while accurately eliminating unstable and less important structures.
    \item  We propose an efficient group-wise fine-tuning strategy to maintain the performance of the LLMs after pruning. It independently quantifies and adjusts the weights for each group, which not only boosts the computational efficiency but also simplifies the deployment process.
\end{itemize}

\section{Related Work}
Recently, many leading companies have released their open-source LLMs, such as LLaMA \cite{ref21}, Vicuna \cite{Vicuna}, and ChatGLM \cite{ref23}, which have significantly influenced the field of natural language processing. Since these models grow in size and complexity, the need for efficient model pruning techniques has become increasingly apparent. Typically, model pruning can be divided into two categories, including structured pruning and unstructured pruning. Structured pruning removes weights according to a predefined network structure. It is particularly beneficial for hardware acceleration because it conforms to the parallelism of modern computing architectures\cite{ref29}. In contrast, unstructured pruning removes weights individually, which often leads to irregular network structures that are difficult to optimize and deploy in practice. In the following, we focus on structured pruning and review its three stages in previous studies, including weight importance estimation, layer-wise pruning, and LLM fine-tuning.

\subsection{Weight Importance Estimation}
LLM-pruner \cite{llm-pruner} was the first framework for structured pruning of LLMs, which effectively removed non-critical coupling structures and sped up the process without relying on the original training data. Following it, LoRAShear \cite{LoRAShear} employed the low-Rank adaptation of LLMs (LoRA) with half-space projected gradient (LHSPG) for progressive pruning, dynamically evaluating weight importance to retain more critical information and achieve superior knowledge transfer. Additionally, SparseGPT \cite{ref31} introduced a second-order pruning method based on weight importance, effectively scaling to GPT models with 10 to 100 billion parameters and significantly enhancing pruning efficiency. Wanda \cite{ref29} offered a new weight importance metric based on weights and activations to improve the pruning performance and speed. Besides, a weight importance-driven non-neural model was proposed in \cite{ref33}, which utilized  gradient boosting decision trees (GBDT) as the accuracy predictor for efficient pruning selection. Furthermore, shortened LLaMA \cite{ref35} adopted deep pruning techniques that integrate weight importance with structural efficiency, achieving comparable performance to width pruning, particularly under memory-constrained scenarios. Despite the achievements, the weight estimation metrics in these works have not accurately calculated the importance of different modules in LLMs. Therefore, they may not work well in cases with large pruning ratios.

\subsection{Layer-wise Pruning}
MINI-LLM \cite{MINI-LLM} proposed a hybrid pruning standard to remove non-critical channels and multi-attention heads by integrating magnitude, activation, and gradient. Subsequently, EDGE-LLM \cite{EDGE-LLM} proposed a layer-wise unified compression method, which achieved layer-by-layer pruning through an adaptive layer adjustment scheme. Furthermore, AlphaPruning \cite{AlphaPruning} utilized the heavy-tailed self-regularization theory to design the layer-wise pruning ratio of LLMs, significantly reducing the mode size while maintaining a reasonable perplexity. Although these studies have delved into the issue of layer-wise pruning ratios, they have not addressed the challenge posed by the significant variance in importance scores across different layers. This disparity hinders the ability to uniformly assess their contributions.

\subsection{LLM Fine-tuning}
Fine-tuning is an important method for enhancing the performance of LLMs in downstream tasks. To address the issues of high computational cost and long training latency associated with standard fine-tuning methods, many parameter-efficient fine-tuning (PEFT) algorithms have been proposed and garnered extensive attention \cite{ref36}. For instance, an adapter method was proposed in \cite{ref37}, which inserted small bottleneck adaptation layers to reduce the number of parameters that need to be updated. In addition, LoRA \cite{ref38} reduced computational overhead by fine-tuning low-rank decompositions within the model. Following it, quantization-aware LoRA (QLoRA) \cite{ref39} enhanced the fine-tuning efficiency and effectiveness by combining quantization with LoRA. While these works can fine-tune LLMs efficiently, existing fine-tuning methods face challenges such as high memory usage and inefficiencies in scaling. However, SAAP streamlines quantization and low-rank adaptation, thereby enhancing deployment efficiency for LLMs across various architectures and scales.
\begin{figure}[t]
    \centering
    \includegraphics[width=0.66\linewidth]{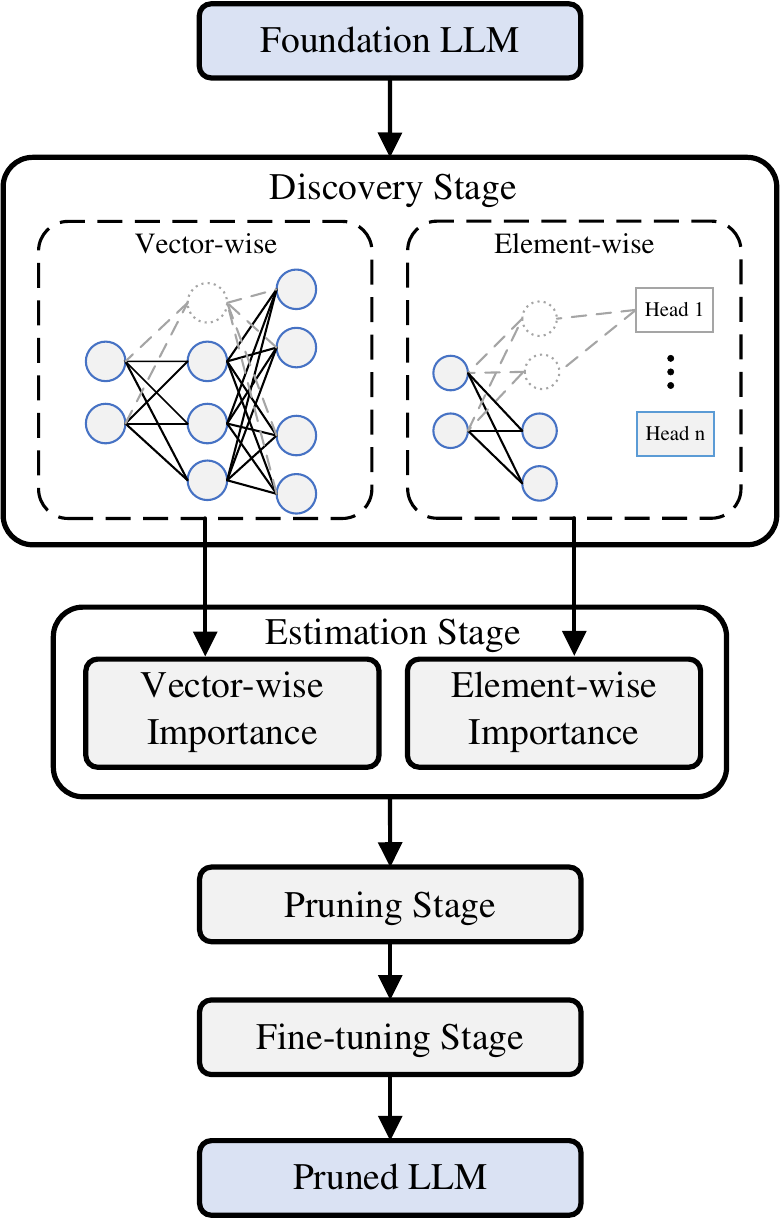}
    \caption{The pipeline of existing LLM pruning methods.}
    \label{fig:LLM_pruner}
\end{figure}
\section{Preliminaries}
\subsection{LLM Pruning Process}
As shown in Fig. \ref{fig:LLM_pruner}, the pruning process of LLMs typically consists of four stages\cite{ref40}: 
\begin{itemize}
    \item \textit{Discovery Stage:} Given a foundation LLM, all coupled structures in the LLM are first identified based on a dependency detection algorithm \cite{ref11}. Each coupled structure is defined as a ``group". 
    \item \textit{Estimation Stage:} When identifying all groups, it is necessary to evaluate the importance of each group. There are two types of importance metrics, including vector-wise importance and element-wise importance. Specifically, let ${\mathbf{W}}_i^{\rm{V}}$ denote the weights of the $i$-th group. Then, the vector-wise importance of group $i$ is given by
    \begin{equation}
    \begin{aligned}
    &\!\!\!\!I_i^{\rm{V}} = \left \vert{\rm{\Delta}}{\cal L}({\cal D}) \right \vert \\
    &\!\!\!\!= \left \vert {{{\cal L}_{{\mathbf{W}}_i^{\rm{V}}}}({\cal D}) \!-\! {{\cal L}_{{\mathbf{W}}_0^{\rm{V}}}}({\cal D})} \right \vert\\
    {\rm{    }} 
    &\!\!\!\!= \left \vert \!
    \frac{{\partial {{\cal L}^{\top}}({\cal D})}}{{\partial {\mathbf{W}}_i^{\rm{V}}}}{\mathbf{W}}_i^{\rm{V}}
    \!-\! \frac{1}{2}({\mathbf{W}}_i^{\rm{V}})^{\top}{\mathbf{H}}{\mathbf{W}}_i^{\rm{V}} \!+\! {\cal O}\left( {||{\mathbf{W}}_i^{\rm{V}}|{|^3}} \right) \!\right \vert,
    \end{aligned}
    \end{equation}
where $\mathcal{L}$ is the next-token prediction loss, $\mathcal{D}$ is the training dataset, $\top$ represents the transpose of the matrix, ${\mathbf{H}}$ is the Hessian matrix of ${\mathbf{W}}_i^{\rm{V}}$. ${\cal O}\left( {\Vert{\mathbf{W}}_i^{\rm{V}}\Vert^3} \right)$ denotes the high-order terms of Taylor expansion, which can be ignored because the redirection value is small and has little impact on the value of the importance. For the element-wise importance, let ${\mathbf{W}}_i^{\rm{E}}$ denote the weights of each element within the weight matrix ${\mathbf{W}_i}$. Then, the element-wise importance can be approximated by 
\begin{equation}
\begin{aligned}
    &I_i^{\rm{E}} = \left \vert {{{\cal L}_{{\mathbf{W}}_i^{\rm{E}}}}({\cal D}) - {{\cal L}_{{\mathbf{W}}_0^{\rm{E}}}}({\cal D})} \right \vert\\
    & \approx \! \left\vert \frac{{\partial {\cal L}({\cal D})}}{{\partial {\mathbf{W}}_i^{\rm{E}}}}{\mathbf{W}}_i^{\rm{E}} \!-\! \frac{1}{2}\mathop \sum \limits^N_{j = 1} {\left( {\frac{{\partial {\cal L}\left( {{{\cal D}_j}} \right)}}{{\partial {\mathbf{W}}_i^{\rm{E}}}}{\mathbf{W}}_i^{\rm{E}}} \right)^2} \!\!\!+\! {\cal O}\left( {{\rm{||}}{\mathbf{W}}_i^{\rm{E}}{\rm{|}}{{\rm{|}}^3}} \right) \! \right\vert,
\end{aligned}
\end{equation}
where $N$ is the number of data samples in the dataset $\mathcal{D}$ and $\mathcal{D}_j$ is the $j$-th data sample.
    \item \emph{Pruning Stage:} After finishing the importance estimation, the importance values of all groups (i.e., $I_i^{\rm{V}}$ or $I_i^{\rm{E}}$) are sorted. The groups with lower importance values are removed based on a predefined pruning ratio.
    \item \emph{Fine-tuning Stage:}  To mitigate the performance degradation caused by pruning, LoRA is adopted to fine-tune the pruned model using a small dataset \cite{ref38}. Given the weight matrix $\mathbf{W}$ is approximated by two low-rank matrices $\mathbf{P}$ and $\mathbf{Q}$, it follows
\begin{equation}
f(x) = (\mathbf{W} + \mathbf{\Delta W}) x + \mathbf{b} = (\mathbf{W} x + \mathbf{b}) + (\mathbf{P} \mathbf{Q}) x,
\end{equation}
where $\Delta \mathbf{W} = \mathbf{P} \mathbf{Q}$ and $\bf{b}$ is the bias term. By fine-tuning $\mathbf{P}$ and $\mathbf{Q}$, we can obtain the pruned LLM with low computational complexity.
\end{itemize}

\subsection{Challenges in LLM Pruning}
Although the aforementioned methods can effectively prune LLMs with little performance degradation, they still face three key challenges:
\begin{itemize}
\item {\bf{Single metric evaluation.}} Existing LLM pruning methods mainly utilize a single metric to evaluate the importance of all groups. However, due to the complex interdependence of LLMs, the evaluation result may be inaccurate, thereby affecting the pruning performance.
\item {\bf{Uniform pruning ratio.}} Most previous works adopt a uniform pruning ratio across all layers of LLMs, disregarding the distinct contributions of different structures. Such a straightforward approach may lead to unstable pruning performance when the pruning ratio is large. 
\item {\bf{High memory cost.}} Existing works typically utilize LoRA for model fine-tuning. Nevertheless, LoRA uses 16-bit floating point numbers (FP16), which results in high memory cost and cannot be applied in resource-constrained scenarios.
\end{itemize}

To address these issues, we propose a novel pruning method, namely SAAP, to adaptively remove non-essential structures based on their importance without introducing significant computational and memory costs. SAAP uses an adaptive metric to prune unstable structures and employs group-wise fine-tuning to ensure the performance of the pruned LLM. In the following, we introduce our SAAP method in detail.  
\begin{figure*}
    \centering
    \includegraphics[width=0.9\linewidth]{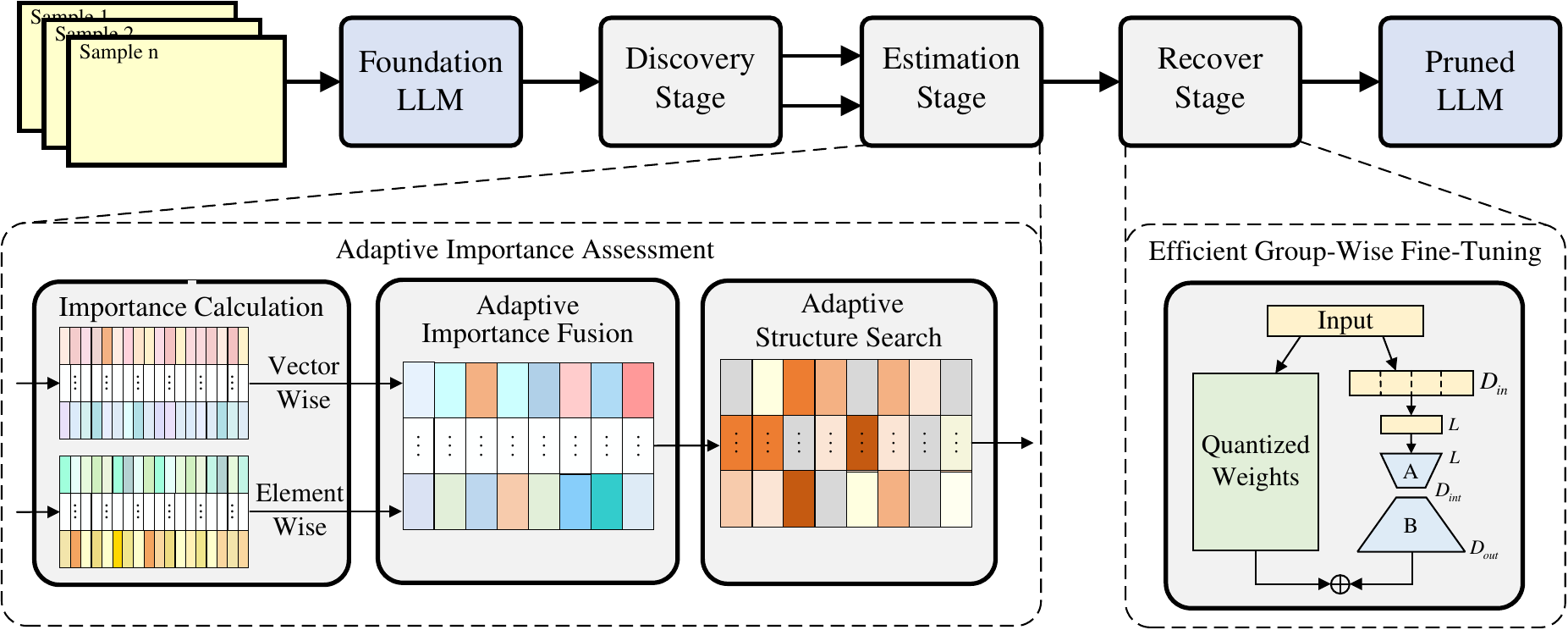}
    \caption{An overview of the SAAP method. Given a foundation LLM, SAAP first removes the most volatile structure by adaptive importance assessment. Then, it restores the performance of the pruned model through efficient group-wise fine-tuning.}
    \label{fig:overview}
\end{figure*}
\section{Method}
In this section, we first provide an overview of the SAAP method. Then, we elaborate on the detailed designs of SAAP, including the adaptive importance assessment approach and the efficient group-wise fine-tuning scheme.
 \subsection{Overview of SAAP}
As illustrated in Fig. \ref{fig:overview}, SAAP follows the structured pruning process consisting of three stages, i.e., discovery stage, estimation stage, and recover stage. The discovery stage identifies all groups in the LLM, while the estimation stage and recover stage evaluate the importance of each group and restore the model performance, respectively. 
The key innovations of SAAP lie in two aspects. 
\begin{itemize}
    \item SAAP introduces an adaptive stability indicator in the estimation stage to assess unstable and redundant components of the network. By combining both coarse-grained and fine-grained information, SAAP better captures the varying significance of different coupled structures and improves the accuracy of importance estimation. 
    \item Furthermore, SAAP extends its approach in the estimation stage by proposing an adaptive structure search strategy. This strategy evaluates the stability of importance scores across different structures, enabling a unified assessment that identifies and prunes unstable coupled structures more effectively.
    \item SAAP employs an efficient group fine-tuning strategy in the recover stage, which maintains the accuracy of the pruned LLM without incurring much computational cost. 
\end{itemize}

\subsection{Adaptive Importance Assessment}
As shown in Fig. \ref{fig:overview}, the estimation stage of SAAP comprises three components, including importance calculation, adaptive importance fusing, and adaptive structure search. The importance calculation is the same as that in LLM-pruner \cite{llm-pruner}. The adaptive importance fusion adaptively combines the coarse-grained and fine-grained information to evaluate the importance of each group. The adaptive structure search calculates a standard indicator of importance fluctuation to facilitate stable pruning LLMs.

\textbf{a) Adaptive importance fusion.} In this work, we develop a multi-task loss function by maximizing the uncertainty in an equal variance Gaussian likelihood. Specifically, let $F({I_W})$ denote the adaptive importance fusion metric with input weight matrix ${\mathbf{W}}$. Then, for regression tasks, the output typically follows a Gaussian distribution. Thus, the probability distribution of the output $y$ can be expressed as
\begin{equation}
{P} \left( {y|F({I_W})} \right) = \mathcal {N}\left( {F({I_W}),{\lambda ^2}} \right),
\end{equation}
where $\lambda$ represents the scalar noise. For classification tasks, we usually convert the model's output into a probability vector using the softmax function, i.e.,
\begin{equation}
P\left( {y|F({I_W})} \right) = {\rm{Softmax}}\left({F({I_W})}\right),
\end{equation}
where $I_W$ refers to the importance calculated in LLM-pruner. Given some sufficient statistics, we define the likelihood function that can be factorized over multiple outputs. Each output depends on the network's sufficient statistics $F({I_W})$, as given by
\begin{equation}
\begin{aligned}
&P \left( {{y_1}, \ldots ,{y_K}|F({I_W})} \right) \\& = P\left( {{y_1}|F({I_W})} \right) ,\ldots, P\left( {{y_K}|F({I_W})} \right)\}.
\end{aligned}
\end{equation}

In the maximum likelihood estimation, we optimize model parameters by maximizing the logarithm of the likelihood function. The logarithm likelihood expression is given by
\begin{equation}
\log P\left( {y|F({I_W})} \right)\propto- \frac{1}{{2{\lambda ^2}}}{| {y - F({I_W})}|^2} - \log \lambda,
\end{equation}
where $\lambda$ represents the observation noise parameter of the model, reflecting the amount of noise in the output. Our goal is to maximize the log-likelihood for model parameters $W$ and noise parameters $\lambda$. In the adaptive importance fusion metric task, the model's output consists of two vectors, ${y_1}$ and ${y_2}$, which represent importance outputs for vector-wise and element-wise in LLM-pruner, respectively. Both vectors follow a Gaussian distribution, i.e.,
\begin{equation}\label{eq.8}
\begin{aligned}
\left( {{y_1},{y_2}|F({I_W})} \right) = P\left( {{y_1}|F(I_i^{\rm{V}})} \right) \cdot P\left( {{y_2}|F(I_i^{\rm{E}})} \right)\\
= {\cal N}\left( {{y_1};F(I_i^{\rm{V}}),{\lambda _1}^2} \right) \cdot {\cal N}\left( {{y_2};F(I_i^{\rm{E}}),{\lambda _2}^2} \right).
\end{aligned}
\end{equation}

We calculate the minimization objective of the model based on \eqref{eq.8}. The adaptive importance score $I_i^{{\rm{ada}}}$ is calculated as 
\begin{equation}
\begin{aligned}
&I_i^{{\rm{ada}}} = - \log P\left( {{y_1},{y_2}|F({I_W})} \right)\\
&\propto \frac{1}{{2{\lambda _1}^2}}{| {{y_1} - F(I_i^{\rm{V}})} |^2} + \frac{1}{{2{\lambda _2}^2}}{| {{y_2} - F(I_i^{\rm{E}})} |^2} + \log {\lambda _1}{\lambda _2}\\
 &= \frac{1}{{2{\lambda _1}^2}}I_i^{\rm{V}} + \frac{1}{{2{\lambda _2}^2}}I_i^{\rm{E}} + \log {\lambda _1}{\lambda _2}.
\end{aligned}
\end{equation}

We define $I_i^{\rm{V}}$ as the coarse-grained importance score, denoted as $\left\| {{y_1} - F(I_i^{\rm{V}})} \right\|^2$. Similarly, $I_i^{\rm{E}}$ is defined as the fine-grained importance score. $I_i^{{\rm{ada}}}$ represents the importance score after adaptive fusion.

\begin{figure}
    \centering
    \includegraphics[width=0.97\linewidth]{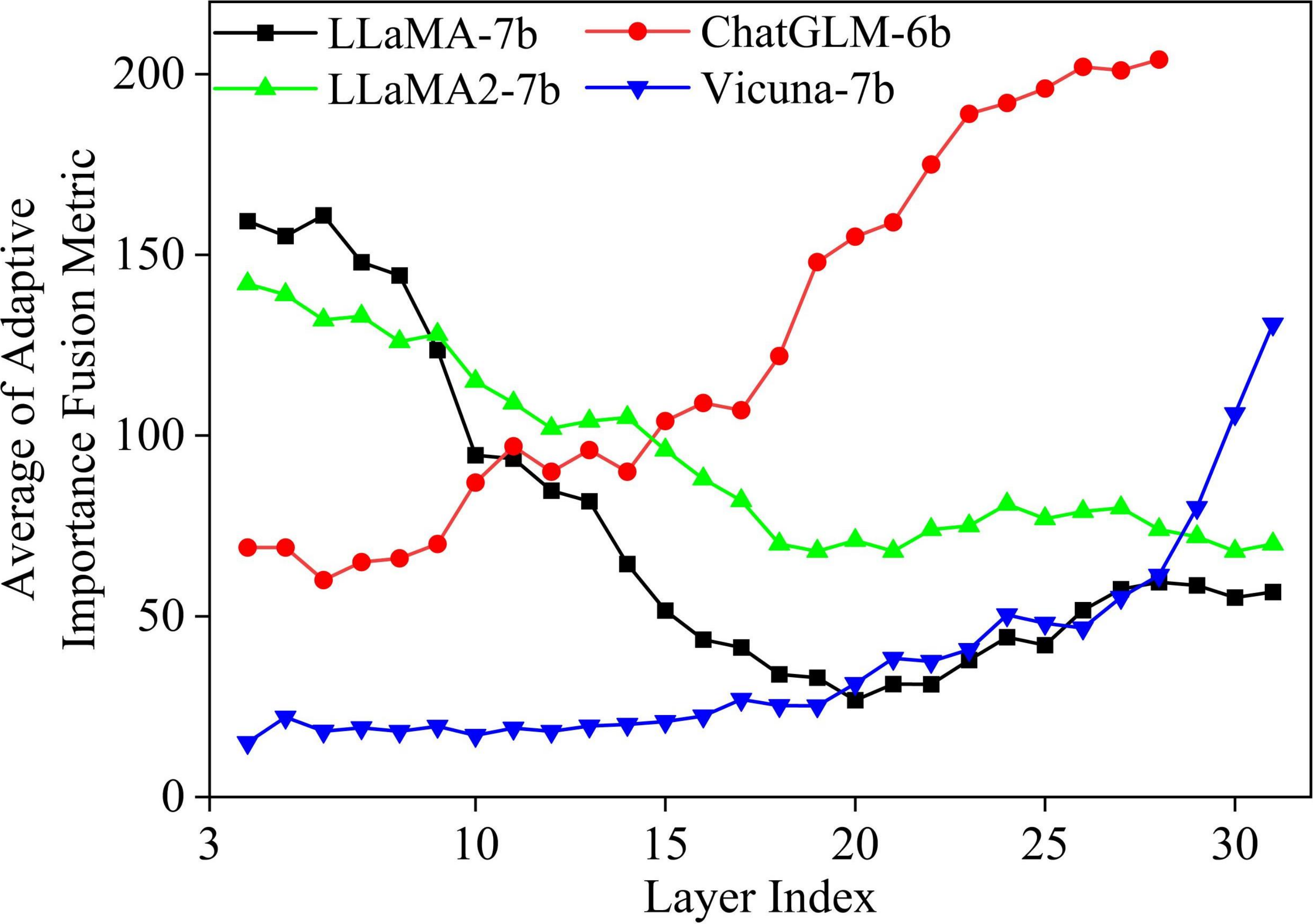}
    \caption{Average of adaptive importance fusion metrics of each layer in different LLMs.}
    \label{fig:importance_fusion}
\end{figure}

\begin{remark}
The adaptive importance assessment realizes the unified calculation of the importance scores of different layers and modules by adaptively fusing the importance information of different weights. This construction is smooth and differentiable, and does not converge to zero. The simple linear weighting may result in the importance being zero, which will affect the calculation of the adaptive structure search part. The adaptive importance fusion metric can be optimized according to the model at different layers and stages. Such a design can flexibly adapt to LLMs with varying network structures and parameter scales.
\end{remark}

\textbf{b) Adaptive structure search.} Structured pruning is primarily based on ``layered pruning." However, different layers and modules have distinct behaviors, as shown in Fig. 3. Hence, it is hard to apply a unified pruning approach \cite{ref44}.

To address this challenge, we introduce the importance fluctuation indicator as a unified measure of importance calculated for each layer or module, i.e.,
\begin{equation}\label{eq.10}
{M_{l,j}} = \frac{1}{{D - 1}}{\sum\limits_{d = 1}^D {(I_{l,j}^d - I_{l,j}^D)} ^2}\left\| {{{\mathbf{W}}_{l,j}}} \right\|_2^2,
\end{equation}
where ${M}_{l,j}$ represents the proposed importance fluctuation indicator. $\left\| {{{\mathbf{W}}_{l,j}}} \right\|_2^2$ denotes the squared norm of the weight coefficients for channel $j$ in layer $l$. $ I$ is the adaptively fused importance score $I_i^{{\rm{ada}}}$ obtained from previous calculations. $I_{l,j}^d $ signifies the importance score for channel $j$ in layer $l$ under calibration samples of $d$, while $I_{l,j}^D$ represents the average importance score for channel $j$ in layer $l$ under calibration samples of $D$. Due to employing Bessel correction \cite{ref45} for unbiased estimation, $\frac{1}{{D - 1}}$ is adopted.

Next, we calculate the adaptive stability indicator, which captures relative changes and is suitable for the final unified search in structured pruning, i.e.,

\begin{equation}
{\hat M_{l,j}} = \frac{{{M_{l,j}} - \mathrm{mean}{\rm{[}}{M_{l,j}}{\rm{]}}}}{{\sqrt {\mathrm{mean}{{[{M_{l,j}} - \mathrm{mean}{\rm{[}}{M_{l,j}}{\rm{]]}}}^2}} }},
\end{equation}
where $\mathrm{mean}{\rm{[}}{M_{l,j}}{\rm{]}}$ is the average value of $M_{l,j}$, with the denominator in the formula representing the calculation of standard deviation. $\hat M_{l,j}$ represents the adaptive stability indicator, which directly reflects the relative volatility of importance scores. Higher relative volatility indicates redundancy and instability within the entire model. Finally, based on the pruning ratio of the model, layers or modules with maximum relative volatility are removed to complete model pruning.

\begin{remark}
The adaptive structure search can effectively capture the importance fluctuations of different layers and modules in the model by introducing a relative stability indicator. This method overcomes the limitations of existing pruning methods in uniformly evaluating the importance of model parameters, and can accurately identify and remove redundant structures that are unstable under different inputs or conditions. This not only improves the effectiveness of the pruning process but also ensures the stability and robustness of the model after pruning.
\end{remark}

Compared with existing LLM pruning methods, the introduction of the adaptive importance fusion metric and the adaptive structure search not only solves the problem of importance score differences between different structural levels but also provides more precise guidance in the overall pruning process.

\subsection{\textit{\textit{Efficient Group-Wise Fine-Tuning}}}
In the recovery stage, we aim to quantify the pruned model weights to minimize GPU usage and ensure the fine-tuned weights remain quantized, thus improving computational deployment efficiency. QLoRA has recently achieved the first goal by quantifying the model weights from FP16 to NF4 during the fine-tuning stage. However, QLoRA shares the same concept as LoRA. QLoRA introduces matrices $\bf{A}$ and $\bf{B}$, which are adjusted while keeping the model weights $\bf{W}$ unchanged, aiming for efficient fine-tuning. We define the size of $\bf{W}$ is ${D_{in}} \times {D_{out} }$. Then, the post-fine-tuned weight $\bf{W'}$ can be represented as
\begin{equation}
{\bf{W'}} = {\bf{W}} + s \cdot {{\bf{A}}{\bf{B}}},
\end{equation}
where $s$ represents the adjustment parameter of the matrix. The dimensions of $\bf{A}$ and $\bf{B}$ are ${D_{in}} \times {D_{int} }$ and ${D_{int}} \times {D_{ out} }$ respectively. Therefore, the dimension of ${\bf{A}} {\bf{B}}$ is the same as $\bf{W}$. However, it can be observed that after quantization, $\bf{W'}$ contains $ s \cdot {\bf{A}}  {\bf{B}}$, which will result in the final weight matrix $\bf{W'}$. Although post-training quantization is possible, it may reduce model accuracy \cite{ref45}.

To address the aforementioned issues and combine and $ s \cdot {\bf{A}} {\bf{B}}$ without using FP16, we propose a grouped fine-tuning strategy. Each group's weights are independently quantized and adjusted during fine-tuning, as illustrated in Fig. 1. Grouped quantization enhances computational efficiency, simplifies deployment, and prevents the accuracy loss typically associated with post-training quantization.

We first divide each column of weight $\bf{W}$ into $L$ groups, where $L$ is set to be a divisor of the number of columns in $\bf{W}$ to ensure balanced grouping. By grouping the weight $\bf{W}$, we can reduce the dimensionality of ${\bf{A}} $ from ${\bf{A}} {\bf{B}}$ to $L$. We usually set $L \ll {D_{in}}$, so the parameter count of decreases from ${D_{in}} \times {D_{int} }$ to $L \times {D_{int}}$.

For each group, we set $a$ and $b$ as the scaling factor and zero-point offset, respectively. Instead of quantizing each column of $\bf{W}$, we use the scaling factor and zero-point offset for quantization, which are defined as
\begin{equation}
\begin{cases} 
a = \dfrac{{\max (\bf{W}) - \min (\bf{W})}}{{{2^N} - 1}},\\ 
b = \min (\bf{W}),
\end{cases} 
\end{equation}
where $N$ is the number of quantization bits, and we set $N=4$ to use int4 for quantization. We use $a$ and $b$ to restore the quantized weights of each group to their original state, with the specific expression as
\begin{equation}
{{\bf{W}}_l} = a_g({{\bf{W}}_g} - b_g),
\end{equation}
where ${\bf{W}}_g$ represents the quantized weight of group $g$,  $a_g$ and $b_g$ represent the scaling factor $a$ and zero-point offset $b$ of group $g$, respectively. Finally, the weights ${\bf{W}}_l$ adjusted by grouping are arranged back into the matrix in the original order to form a complete fine-tuned weight matrix $\bf{W}$.

By introducing the grouping operation, we reduce the number of quantization parameters from $({D_{in}} \times {D_{int}}  + {D_{int}} \times {D_{out}})$ to $(L \times {D_{int }} + {D_{int}} \times {D_{out}})$, and combine quantization and low rank well.
\begin{remark}
The design of efficient group-wise fine-tuning enables flexible adaptation to models of various sizes, ensuring that the computational load remains balanced while preserving accuracy. This construction simplifies the integration of quantization and low-rank adaptation, enhancing deployment efficiency for LLMs of diverse architectures and scales.
\end{remark}

\section{Experiments}
\subsection{Experimental Settings}
\begin{table*}[ht]
  \centering
  \renewcommand{\arraystretch}{1.25}
  \caption{Zero-Shot Performance of the Compressed LLaMA-7B. The Accuracy Average  is Calculated among the Different Classification Datasets.}
  \label{tab: big_exp}\begin{tabular}{>{\centering\arraybackslash}p{1.3cm}>{\centering\arraybackslash}p{1.45cm}>{\centering\arraybackslash}p{0.4cm}*{9}{c}} 
  \hline
    {\begin{minipage}{2cm}Pruning Ratio\end{minipage}} & Method& PTB↓& WikiText2↓&ARC-e&ARC-c&BoolQ&HellaSwag& PIQA&WinoGrande&OBQA& Accuracy Average↑ \\
    \hline
    \multirow{2}{*}{\begin{minipage}{1.3cm}Ratio=0\%\end{minipage}}     & LLaMA-7B   & - & -    & 72.8  & 47.6  & 76.5  & 76.1      & 79.8  & 70.1   & 57.2  & 68.58   \\
                  & LLaMA-7B*  & 22.14  & 12.62       & 67.45 & 41.38 & 73.18 & 72.99     & 78.35 & 67.01      & 42.4  & 63.25\\
    \hline
    \multirow{5}{*}{\begin{minipage}{1.3cm}Ratio=20\% \\w/o tune\end{minipage}}   & LLM-pruner & 34.21  & 19.09       & 60.94 & 36.52 & 57.06 & \underline{66.80}      & 75.68 & 59.83      & \underline{40.00}    & 56.69              \\
          & LoraPrune  & \underline{34.12}  & 20.67       & \underline{62.14} & 34.59 & 57.98 & 65.81     & 75.11 & 59.9       & 39.98 & 56.5\\
                  & Wanda & 38.19  & 22.12       & 56.63 & 33.98 & \underline{64.93} & 58.12     & 70.14 & 55.39      & 35.43 & 53.5               \\
                  & LoRAShear  & -      & -           & -     & -     & -     & -         & -     & -          & -     & -                  \\
                  & SAAP       & 34.15  & \underline{18.73} & 62.06 & \underline{37.82} & 63.07 & 66.45     & \underline{76.73} & \underline{60.57}      & 39.35 & \underline{58.01}              \\
                  \hline
    \multirow{5}{*}{\begin{minipage}{1.3cm}Ratio=20\%\\ w/ tune\end{minipage}}   & LLM-pruner & 30.11  & 17.58       & 64.31 & 36.77 & 64.62 & 68.8      & 77.2  & 63.14      & 39    & 59.12              \\
          & LoraPrune  & 28.75  & 16.8        & 65.87 & 37.69 & 65.62 &\textbf{ 70.00} & \textbf{79.31} & 62.76  & 39.14 & 60.05              \\
                  & Wanda      & 33.16  & 18.43       & 60.65 & 36.26 & 65.75 & 64.52     & 74.7  & 59.35      & 39.4  & 57.23  \\
                  & LoRAShear  & -      & -   & 64.11 & 38.77 & 70.17 & 68.69     & 76.89 & \textbf{65.83} & \textbf{40.78} & 60.75\\
                  & SAAP  & \textbf{26.3}  & \textbf{14.58}  & \textbf{66.15} & \textbf{39.72} & \textbf{70.28} & 69.82     & 77.26 & 65.29      & 40.55 & \textbf{61.29}\\
                  \hline
    \multirow{5}{*}{\begin{minipage}{1.3cm}Ratio=50\%\\ w/o tune\end{minipage}}    & LLM-pruner & 255.38 & \underline{112.44} & 33.50  & 27.22 & 52.32 & 35.64     & 59.63 & 53.20       & 33.40  & 42.13              \\
         & LoraPrune  & 260.14 & 121.96      & 33.82 & 26.93 & 51.78 & \underline{36.76}     & 56.90  & 53.80       & 33.10  & 41.87 \\
                  & Wanda      & 437.71 & 223.46      & \underline{39.43} & 25.76 & 45.13 & 31.37     & 55.54 & \underline{55.87}      & 30.12 & 40.46 \\
                  & LoRAShear  & -      & -           & -     & -     & -     & -         & -     & -          & -     & -                  \\
                  & SAAP       & \underline{249.54} & 113.27  & 38.36 & \underline{27.52} & \underline{51.93} & 36.51   & \underline{60.38} & 54.69  & \underline{33.51} & \underline{43.27}\\
                  \hline
    \multirow{5}{*}{\begin{minipage}{1.3cm}Ratio=50\%\\w/ tune \end{minipage}}    & LLM-pruner & 66.35  & 38.12       & 45.96 & 29.18 & 60.28 & 47.06     & 69.31 & 53.43      & 35.6  & 48.69              \\
           & LoraPrune  & 50.3   & 30.12       & 45.13 & 31.62 & 61.88 & 47.86     & 71.53 & 55.01      & 34.98 & 49.72              \\
                  & Wanda      & 85.87  & 43.89 & 42.68 & \textbf{34.20}  & 50.9  & 38.12     & 57.38 & 55.98      & 38.78 & 45.43              \\
                  & LoRAShear  & -      & - & 47.68 & 32.26 & 62.12 & 48.01     & 71.80  & \textbf{56.29} & 34.61 & 50.39  \\
                  & SAAP & \textbf{52.58}  & \textbf{29.35} & \textbf{48.13} & 34.11 & \textbf{62.71} & \textbf{48.63} & \textbf{72.08} & 56.12& \textbf{36.58} & \textbf{51.19 } \\
                  \hline
    \end{tabular}
\end{table*}
\textbf{Foundation LLMs.} 
We first select four types of LLaMA \cite{ref21} for experiments, including LLaMA-7B, LLaMA-13B, LLaMA-33B, and LLaMA-65B. These models represent a wide range of computational complexities and capacities, making them suitable for validating the scalability of the proposed SAAP method. Moreover, we conduct comparative analysis on five LLMs, including Vicuna-7B, Vicuna-13B  \cite{Vicuna}, LLaMA2-7B \cite{LLaMA2}, LLaMA2-13B, and LLaMA3-8B \cite{ref48}, demonstrating the versatility of SAAP across different model architectures. 

\textbf{Datasets}. To validate the effectiveness of SAAP, we conduct experiments on nine open-source datasets with two tasks of common sense reasoning and interactive understanding. The ARC Easy dataset and ARC Challenge dataset cover simple and complex scientific questions, respectively \cite{ref49}. The BoolQ dataset \cite{ref50} tests the model's ability to understand complex contexts and perform text extraction. The HellaSwag dataset \cite{ref51} focuses on the model's capability to understand and reason in daily scenarios. The PIQA dataset \cite{ref52} evaluates common sense reasoning, and the WinoGrande dataset \cite{ref53} concentrates on common sense reasoning and contextual understanding. The OBQA dataset \cite{ref54} aims to evaluate and enhance question-answering systems, testing the LLM's broad common sense and multi-step reasoning capabilities. Additionally, we test the zero-shot perplexity (PPL) on the PTB \cite{PTB} and the WikiText2 \cite{ref56} datasets.

\textbf{Baseline methods.} We consider four baseline methods for comparative experiments: 1) LLM-pruner \cite{llm-pruner}, which automatically calculates each group's contribution to model performance and performs effective pruning afterwards.
2) LoraPrune \cite{ref59}, which combines low-rank decomposition with pruning techniques, primarily reducing model parameters through low-rank approximation. 
3) Wanda, which employs an importance metric based on weights and activation values to guide the pruning process.
and 4) LoRAShear \cite{LoRAShear}, which applies the half-space projected gradient (LHSPG) technique to gradually reduce the number of model parameters while preserving the model's ability to transfer knowledge.

\textbf{Performance metrics.} For classification tasks on datasets---ARC, BoolQ, HellaSwag, PIQA, WinoGrande, and OBQA, we utilize the classification accuracy as the performance metric. It is defined as the proportion of correct predictions made by LLMs and measures the generalization ability of LLMs in multi-domain tasks. For language modeling tasks on datasets---PTB and WikiText2, we use perplexity as the performance metric, showcasing the predictive ability of the model. Lower perplexity indicates more accurate next-word predictions by LLMs. We note that PPL is an important indicator for measuring model quality in sequential tasks. Additionally, the inference speed is measured by the number of tokens generated per second. 

\textbf{Implementation details.} Our experiments are conducted on CUDA 12.1 with HuggingFace 4.39.1 and PyTorch 2.2. The experimental platform is Ubuntu 20.04 equipped with two A100 GPUs, each with 80GB of memory. During pruning, we randomly select 50 samples (sequence length $=128$) from the Bookcorpus dataset \cite{ref57} as calibration samples. Moreover, we use the Alpaca dataset \cite{ref58} in the recover stage, which contains 50k samples in total. We set the parameter $L = 32$ for efficient group-wise fine-tuning.

We note that the first three layers and the last layer of LLMs have a significant impact on the model performance. Therefore, we keep them fixed and only prune other layers. Taking LLaMA-7B as an example, if the overall pruning ratio is set to 20\%, we increase the pruning ratio to 25\% specifically for the fourth to 30th layers. In the recover stage, we set the learning rate to $1 \times 10^{-4}$, the warming step to $1,000$, and the batch size to 128. Besides, we use the AdamW optimizer in the experiment.

\begin{table}
  \centering
  \renewcommand{\arraystretch}{1.25}
  \caption{Zero-shot Performance of the LLaMA Model Family on the WikiText-2 Validation set, Measured in Terms of Perplexity}
  \label{tab: different_llama}\begin{tabular}{*{6}{c}}
  \hline
    \multirow{2}{*}{Pruning Ratio} & \multirow{2}{*}{Method}     & \multicolumn{4}{c}{LLaMA}    \\ 
    \cline{3-6}
                  &            & 7B    & 13B   & 33B   & 65B   \\
                  \hline
    0\%           &     -       & 12.62 & 10.81 & 9.11  & 8.21  \\
    \hline
    \multirow{2}{*}{20\% }         & LLM-pruner & 17.58 & 15.18 & -     & -     \\
                  & SAAP       & 14.58 & 13.61 & 12.75 & 11.63  \\
                  \hline
    \multirow{2}{*}{50\%}          & LLM-pruner & 38.12 & -     & -     & -     \\
                  & SAAP       & 32.4  & 24.33 & 22.17 & 18.32 \\
                  \hline
    \end{tabular}
\end{table}

\begin{table}
  \centering
  \renewcommand{\arraystretch}{1.25}
  \caption{Statistics of Inference Speed and Memory Footprint}
  \label{tab: test_token}\begin{tabular}{*{5}{c}}
  \hline
Pruning Ratio & Method     & Memory     & Params & Tokens/s      \\
\hline
0\%           & LLaMA-7B   & 12884.5MiB & 6.74B  & 25.84         \\
\hline
\multirow{2}{*}{20\% } & LLM-pruner & 10375.5MiB & 5.42B  & 32.57(↑27\%) \\
              & SAAP       & 10055.7MiB & 5.26B  & 34.15(↑32\%) \\
              \hline
\multirow{2}{*}{50\%}& LLM-pruner & 6533.9MiB  & 3.35B  & 40.95(↑58\%) \\
              & SAAP       & 5940.8MiB  & 3.12B  & 42.72(↑65\%)\\
              \hline
    \end{tabular}
\end{table}
\begin{figure*}[t]
    \centering
    \renewcommand{\arraystretch}{1.25}
    \includegraphics[width=0.99\linewidth]{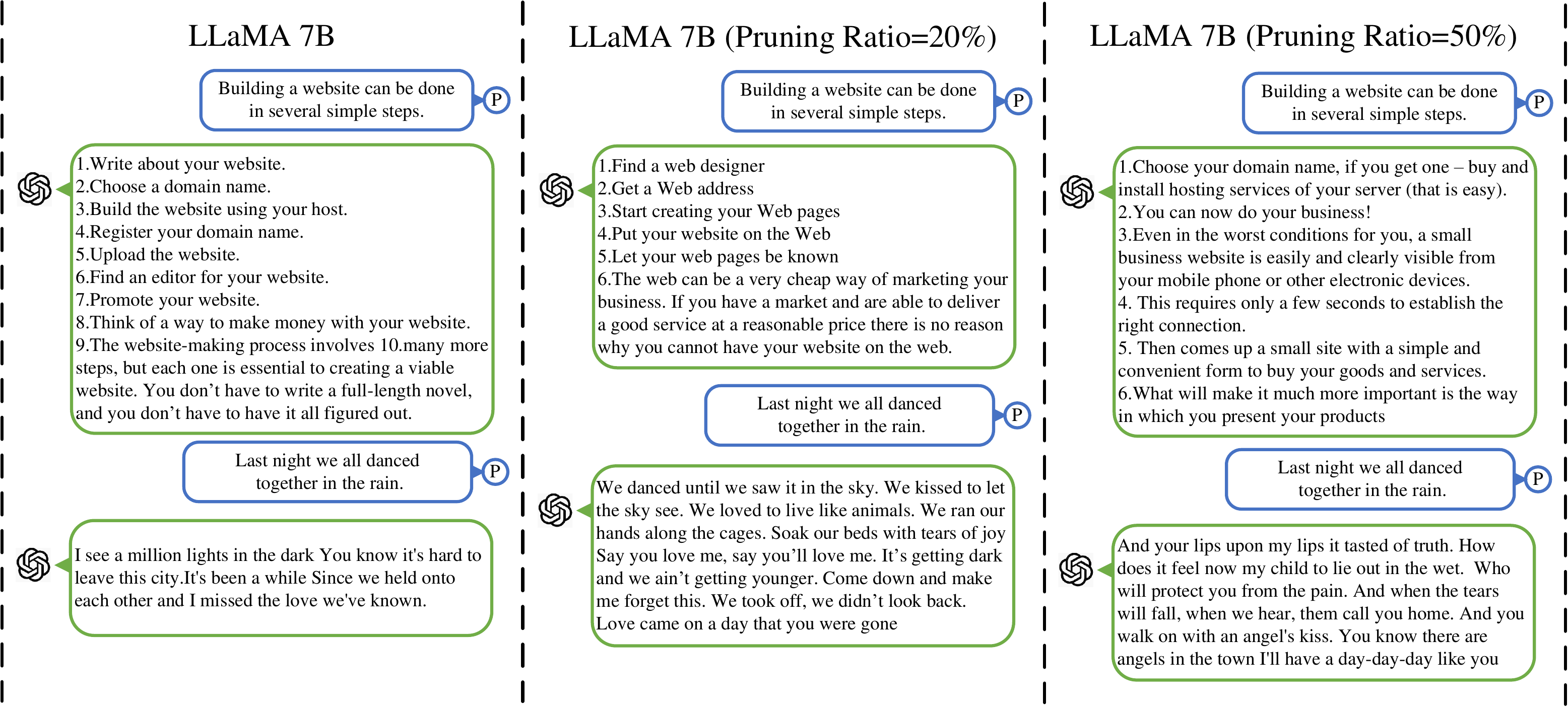}
    \caption{LLM's answer under different pruning ratios.}
    \label{fig:LLM_answer}
\end{figure*}

\begin{table*}[t]
  \centering
  \renewcommand{\arraystretch}{1.25}
  \caption{Zero-Shot Performance of the Compressed Vicuna-7B}
  \label{tab: vicuna-7b}\begin{tabular}{>{\centering\arraybackslash}p{1.3cm}>{\centering\arraybackslash}p{1.45cm}>{\centering\arraybackslash}p{0.8cm}*{7}{c}} 
  \hline
   {\begin{minipage}{2cm}Pruning Ratio\end{minipage}} & Method&ARC-e &ARC-c&BoolQ&HellaSwag& PIQA&WinoGrande&OBQA& Accuracy Average↑ \\
  \hline
    Ratio=0\%     & Vicuna-7B & 65.11 & 41.21 & 76.57 & 70.64     & 77.75 & 67.4       & 40.8  & 62.78              \\   \hline
    Ratio=20\%    & LLM-pruner & 60.98 & 37.12 & 61.7  & \underline{64} & \underline{75.41} & 58.41 & \underline{39} & 56.83              \\
    w/o tune      & SAAP & \underline{61.33} & \underline{37.5}& \underline{65.92} & 63.55     & 74.58 & \underline{60.03} & 38.51 & \underline{57.34}   \\ \hline
    Ratio=20\%    & LLM-pruner & 63.05 & 37.71 & 63.33 & 65.45     & 75.63 & 63.22      & \textbf{39.4}  & 57.78              \\
    w/ tune       & SAAP & \textbf{64.13} & \textbf{39.27} & \textbf{69.15} & \textbf{67.34}  & \textbf{76.02} & \textbf{66.21}  & 38.93 & \textbf{60.15}              \\ \hline
    Ratio=50\%    & LLM-pruner & \underline{33.29} & 27.3  & 53.76 & 34.86     & 59.79 & 50.28 & \underline{34.6}  & 41.98              \\
    w/o tune      & SAAP & 32.95 & \underline{29.18} & \underline{55.56} & \underline{41.26}   & \underline{60.11} & \underline{53.15} & 31.28 & \underline{43.35}    \\ \hline
    Ratio=50\%    & LLM-pruner  & 46.89 & 29.01 & 58.87 & 46.38     & 69.48 & 54.78  &  \textbf{34.8}  & 48.6               \\
    w/ tune  & SAAP  &  \textbf{48.27} &  \textbf{32.47} &  \textbf{59.12} &  \textbf{51.6} &  \textbf{71.93} &  \textbf{55.06}  & 34.17 & \textbf{50.37}   \\          \hline
    \end{tabular}
\end{table*}

\begin{table*}[ht]
  \centering
  \renewcommand{\arraystretch}{1.25}
  \caption{Zero-Shot Performance of the Compressed LLaMA-13B}
  \label{tab: llama13b}\begin{tabular}{>{\centering\arraybackslash}p{1.3cm}>{\centering\arraybackslash}p{1.65cm}>{\centering\arraybackslash}p{0.8cm}*{7}{c}} 
  \hline
   {\begin{minipage}{2cm}Pruning Ratio\end{minipage}} & Method& ARC-e&ARC-c&BoolQ&HellaSwag& PIQA&WinoGrande&OBQA& Accuracy Average↑ \\
  \hline
  Ratio=0\%     & LLaMA-13B & 74.58 & 44.54 & 68.47 & 76.24     & 78.89 & 70.09      & 42    & 64.97              \\ \hline
Ratio=20\%    & LLM-pruner & \underline{64.44} & 36.26 & 63.33 & 63.54     & 73.18 & 60.85      & 38    & 57.09              \\
w/o tune      & SAAP  & 64.1  & \underline{38.27} & \underline{68.25} & \underline{68.17}  & \underline{73.51} & \underline{63.05} & \underline{38.1}  & \underline{59.06}  \\ \hline
Ratio=20\%    & LLM-pruner   & 62.08 & 38.99 & \textbf{69.2}  & 68.89     & 76.55 & 66.38      & 39.6  & 60.24              \\
w/ tune       & SAAP  & \textbf{68.36} & \textbf{41.62} & 68.57 & \textbf{73.81}     & \textbf{77.02} & \textbf{68.32}  & \textbf{40.75} & \textbf{62.63} \\ \hline
Ratio=50\%    & LLM-pruner & 38.93 & 30.03 & \underline{61.83} & 45.49     & \underline{67.08} & 52.09      & \underline{33}    & 46.92              \\
w/o tune & SAAP & \underline{41.62} & \underline{32.55} & 59.34&\underline{50.61}     & 65.52 & \underline{57.33}      & 31.74 & \underline{48.39 }   \\ \hline
Ratio=50\%    & LLM-pruner  & 56.86 & 32    & 62.17 & 57.3      & \textbf{72.85} & 56.99      & \textbf{38.4}  & 53.8               \\
w/ tune & SAAP & \textbf{59.42} & \textbf{36.62} & \textbf{62.25} &\textbf{60.87}    & 72.31 & \textbf{58.17}  & 36.28 & \textbf{55.13}     \\ \hline
    \end{tabular}
\end{table*}

\subsection{\textit{\textit{Performance Comparison with Baseline Methods}}}
In the model pruning process, we use 50 randomly-selected samples from the Bookcorpus dataset \cite{ref57} to estimate performance metrics in our method. We measure the post-pruning performance of the model through perplexity and average accuracy. Table \ref{tab: big_exp} shows the performance comparison of our SAAP method with the four baseline methods at different pruning ratios under LLaMA-7B. The underline (`\_') indicates the best performance achieved solely through pruning, while `bold' denotes the best performance achieved through post-training. Results marked with (*) are below the official results, as some metrics are not provided in \cite{ref21}.

Table \ref{tab: big_exp} demonstrates the effectiveness of the proposed SAAP method. Without fine-tuning, our method achieves the optimal performance, with an average accuracy of 61.3\% across multiple inference datasets at a 20\% pruning ratio. After fine-tuning, our method outperforms existing structured pruning approaches for LLMs. At a 50\% pruning ratio, it achieves the highest average accuracy and lower perplexity. Moreover, our method effectively retains the generalization capabilities of LLMs at high pruning ratios, outperforming other baseline methods. It is observed that at a 20\% pruning ratio, SAAP outperforms the second best method by 1.32\% (without fine-tuning) and 0.54\% (after fine-tuning). At a 50\% pruning ratio, SAAP outperforms the second best method by 1.14\% (without fine-tuning) and 0.8\% (after fine-tuning), showcasing a more evident effect at a higher pruning ratio.

We then conduct experiments on LLaMA models with varying parameter sizes to assess the effectiveness of our proposed method. Table \ref{tab: different_llama} displays the performance for two different pruning ratios across models with 7B, 13B, 33B, and 65B parameters. Similar to the previous experiment, we use 50 randomly selected samples from the Bookcorpus dataset for the SAAP calculations during the estimation stage. The performance of the proposed method is further validated on the WikiText2 test set. It can be seen that SAAP has better performance at both 20\% and 50\% pruning ratios. These results confirm the superior performance and efficacy of our pruning approach. Moreover, we perform language generation tests on the LLaMA-7B model at various pruning ratios, and the results are shown in Fig. \ref{fig:LLM_answer}. This also proves that SAAP performs better both before and after fine-tuning. It is seen that the performance of SAAP  is relatively reasonable and similar to the model without pruning after 20\% and 50\% pruning.

Structured pruning offers better hardware compatibility and deployment convenience than unstructured pruning, making it a more commonly used model compression technique. Table \ref{tab: test_token} presents statistical data from our experiments on the 7B model, including parameter count, memory requirements, and tokens per second. Conducting these tests on a single RTX3090 using the WikiText2 test set, our method demonstrates significant efficiency improvements, reduced parameter count, and faster inference speeds.

In addition to the primary experiments on the LLaMA and Vicuna models, further evaluations are conducted to assess the generalization and robustness of the proposed SAAP method across different LLMs and various pruning scenarios. The next subsection details the results of these extended experiments, highlighting the performance of SAAP in comparison to baseline methods, specifically LLM-pruner and LoRAShear, on a variety of models, including LLaMA2-7B, LLaMA2-13B, and LLaMA3-8B.

\begin{figure*}[t]
    \centering
    \renewcommand{\arraystretch}{1.3}   
    \includegraphics[width=0.85\linewidth]{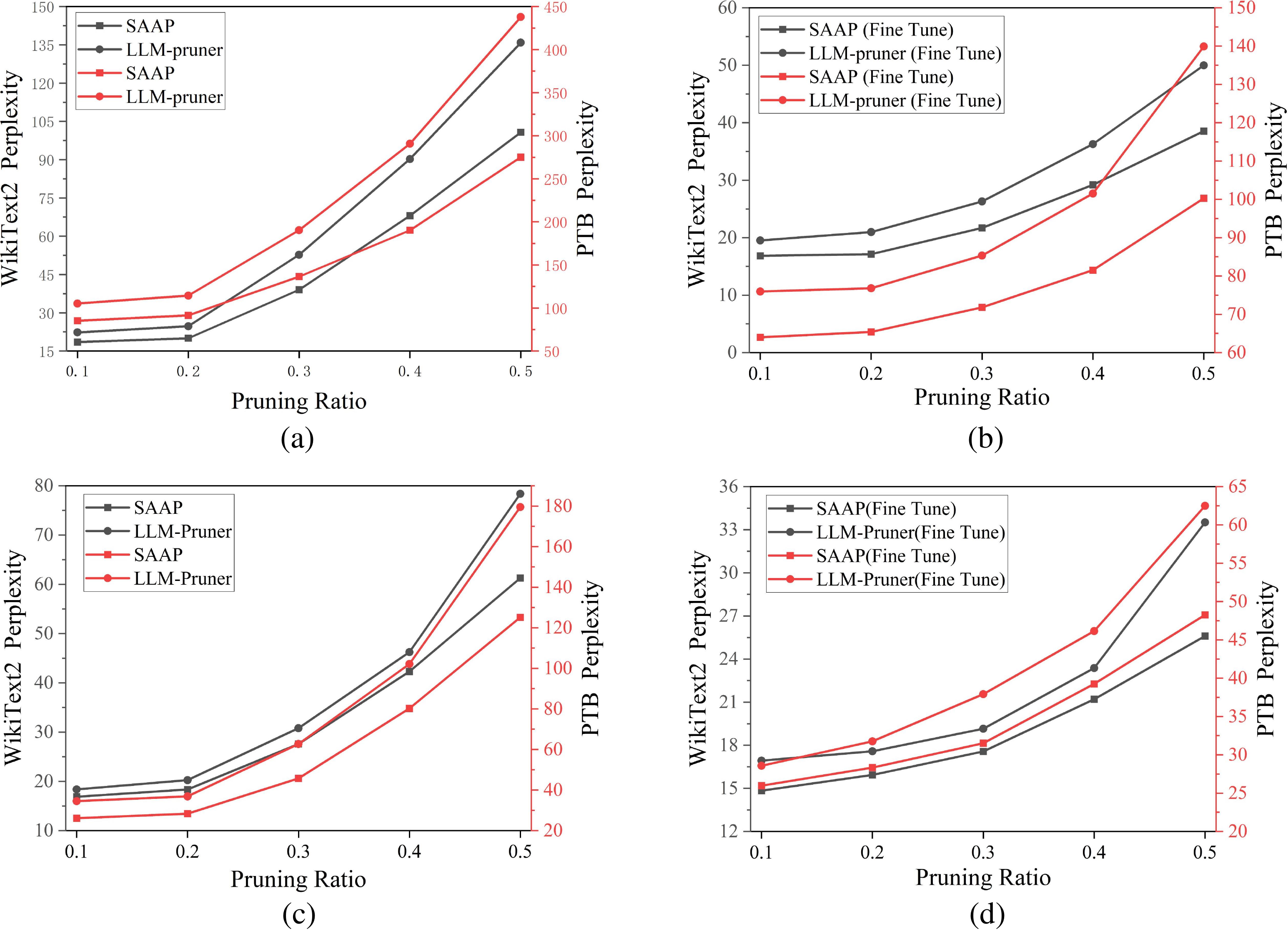}
    \caption{The results of SAAP and LLM-pruner at different pruning ratios. (a) and (b) show the results of the Vicuna-7B model on the PTB and WikiText2 datasets, respectively. (c) and (d) show the results of the LLaMA-13B model on the PTB and WikiText2 datasets, respectively.}
    \label{fig:result}
\end{figure*}
\subsection{Generalization Experiments}
We first conduct comparative experiments between the LLM-pruner and SAAP on the Vicuna-7B and LLaMA-13B models,  with pruning ratios set to 20\% and 50\%, respectively.  The results show that SAAP outperforms LLM-pruner at both pruning ratios. Compared with LLM-pruner, with a 20\% pruning ratio, SAAP demonstrates significant advantages in both accuracy and inference speed. At a 50\% pruning ratio, SAAP maintains high inference performance while keeping the model complexity low. The results of Vicuna-7B and LLaMA-13B on the PTB and Wikitext2 datasets at different pruning ratios are shown in Fig. \ref{fig:result}, Table \ref{tab: vicuna-7b}, and Table \ref{tab: llama13b}.

To further test the generalization ability of SAAP, we conduct experiments not only on different parameter sizes of LLaMA and LLaMA2 but also on the latest LLaMA3-8B model. The results confirm that SAAP is effective not only on earlier versions of LLaMA but also on newer LLM architectures. The detailed results are shown in Table \ref{tab: llama2-7b}, Table \ref{tab: llama2-13b}, and Table \ref{tab: llama3-8b}, which demonstrate the effectiveness of SAAP in different versions of LLaMA. 

We also test SAAP on the Vicuna-7B and 13B models, with specific results displayed in Table \ref{tab: vicuna-7b} and Table \ref{tab: vicuna13b}. These experiments demonstrate that SAAP achieves optimal results at both 20\% and 50\% pruning ratios, further confirming its applicability and generalizability across different LLM architectures and parameter scales.
\begin{table*}
  \centering
  \renewcommand{\arraystretch}{1.25}
  \caption{Zero-Shot Performance of the Compressed Vicuna-13B}
  \label{tab: vicuna13b}\begin{tabular}{>{\centering\arraybackslash}p{1.45cm}>{\centering\arraybackslash}p{1.45cm}>{\centering\arraybackslash}p{0.4cm}*{9}{c}} 
  \hline
   Method & {\begin{minipage}{2cm}Pruning Ratio\end{minipage}}& PTB↓& WikiText2↓&ARC-e&ARC-c&BoolQ&HellaSwag& PIQA&WinoGrande&OBQA& Accuracy Average↑ \\
  \hline
    Base     & Ratio=0\%     & 56.43  & 13.51       & 72.35 & 44.8  & 76.51 & 74.63     & 78.73 & 69.06      & 41    & 65.3               \\ \hline
    \multirow{2}{*}{w/o tune} & Ratio=20\%    & 80.81  & 19.14       & 63.98 & 39.16 & 70.23 & 66.54     & 72.96 & 61.56      & 39.6  & 59.15              \\
             & Ratio=50\%    & 241.35 & 75.64       & 37.25 & 32.61 & 62.81 & 46.13     & 65.07 & 54.29      & 33.72 & 48.27   \\ \hline
    \multirow{2}{*}{w/ tune}  & Ratio=20\%    & 66.52  & 17.08       & 70.58 & 40.87 & 74.45 & 71.07     & 77.09 & 65.9       & 40.8  & 62.97              \\
             & Ratio=50\%    & 85.31  & 26.48       & 62.05 & 32.79 & 68.3  & 60.81     & 72.63 & 57.33      & 38.17 & 56.01  \\ \hline
    \end{tabular}
\end{table*}

\begin{table*}
  \centering
  \renewcommand{\arraystretch}{1.25}
  \caption{Zero-Shot Performance of the Compressed LLaMA2-7B}
  \label{tab: llama2-7b}\begin{tabular}{*{10}{c}} 
  \hline
  Method   & Pruning Ratio & ARC-e & ARC-c & BoolQ & HellaSwag & PIQA  & WinoGrande & OBQA  & Accuracy Average↑ \\ \hline
Base     & Ratio=0\%     & 75.2  & 45.9  & 77.4  & 77.2      & 78.8  & 69.2       & 58.6  & 68.9              \\ \hline
\multirow{2}{*}{w/o tune} & Ratio=20\%    & 72.35 & 40.29 & 68.42 & 71.08     & 72.65 & 61.07      & 52.37 & 62.6              \\
         & Ratio=50\%    & 49.51 & 32.2  & 58.16 & 50.38     & 60.35 & 56.83      & 45.13 & 50.36             \\\hline
\multirow{2}{*}{w/ tune}  & Ratio=20\%    & 74.69 & 44.23 & 73.57 & 74.39     & 76.61 & 68.14      & 55.04 & 66.67             \\
         & Ratio=50\%    & 58.49 & 39.54 & 63.07 & 59.1      & 68.23 & 59.05      & 49.37 & 56.69            \\ \hline
    \end{tabular}
\end{table*}

\begin{table*}
  \centering
  \renewcommand{\arraystretch}{1.25}
  \caption{Zero-Shot Performance of the Compressed LLaMA2-13B}
  \label{tab: llama2-13b}\begin{tabular}{*{10}{c}} 
  \hline
  Method   & Pruning Ratio & ARC-e & ARC-c & BoolQ & HellaSwag & PIQA  & WinoGrande & OBQA  & Accuracy Average↑ \\ \hline
Base     & Ratio=0\%     & 77.30  & 49.4  & 81.7  & 80.70      & 80.50  & 72.8       & 57    & 71.34              \\ \hline
\multirow{2}{*}{w/o tune} & Ratio=20\%    & 71.46 & 43.81 & 76.51 & 74.39     & 76.68 & 67.2       & 54.61 & 66.38      \\
         & Ratio=50\%    & 45.29 & 35.11 & 64.53 & 57.84     & 65.71 & 59.38      & 49.53 & 53.01 \\ \hline
\multirow{2}{*}{w/ tune}  & Ratio=20\%   & 75.18 & 44.92 & 80.03 & 79.15     & 77.84 & 70.02      & 56.24 & 69.05    \\
         & Ratio=50\%    & 53.05 & 39.63 & 71.75 & 65.41     & 71.47 & 64.52      & 51.38 & 59.6  \\ \hline
    \end{tabular}
\end{table*}

\begin{table*}
  \centering
  \renewcommand{\arraystretch}{1.25}
  \caption{Zero-Shot Performance of the Compressed LLaMA3-8B}
  \label{tab: llama3-8b}\begin{tabular}{*{10}{c}} 
  \hline
  Method   & Pruning Ratio & PTB↓  & WikiText2↓ & ARC-e & ARC-c & HellaSwag & PIQA  & WinoGrande & Accuracy Average↑ \\ \hline
Base     & Ratio=0\%     & 10.6  & 6.1        & 80.1  & 50.4  & 60.2      & 79.9  & 72.8       & 68.68 \\ \hline
\multirow{2}{*}{w/o tune} & Ratio=20\% & 19.35  & 10.69      & 74.35 & 46.1  & 55.83     & 75.34 & 65.19      & 63.32 \\
         & Ratio=50\%    & 214.71 & 98.79      & 49.35 & 35.84 & 32.76     & 61.26 & 57.6       & 47.36 \\ \hline
\multirow{2}{*}{w/ tune}  & Ratio=20\%   & 12.76  & 8.35       & 78.13 & 49.75 & 56.35     & 78.02 & 70.33      & 66.52 \\
         & Ratio=50\%    & 39.75  & 25.83      & 59.65 & 42.02 & 46.26     & 73.58 & 62.25      & 56.75 \\ \hline
    \end{tabular}
\end{table*}

\begin{table}
\centering
  \renewcommand{\arraystretch}{1.25}
  \caption{Ablation Study for Adaptive Importance Fusion Metric}
  \label{tab: ablation_importance_estimation}\begin{tabular}{*{3}{c}}
  \hline
Pruning Ratio & Pruning Metric & WikiText2↓ \\
\hline
\multirow{3}{*}{20\%}& Separate Cal   & 15.82       \\
              & Weight Fusion  & 16.35       \\
              & SAAP           & 14.58       \\
              \hline
\multirow{3}{*}{50\%}& Separate Cal   & 31.73       \\
              & Weight Fusion  & 32.41       \\
              & SAAP           & 29.35   \\
              \hline
    \end{tabular}
\end{table}

\begin{table}
\centering
  \renewcommand{\arraystretch}{1.25}
  \caption{Ablation Study for Adaptive Stability Indicator}
  \label{tab: ablation_ASI}\begin{tabular}{*{3}{c}}
  \hline
Pruning Ratio & Pruning Metric & WikiText2↓ \\
\hline
\multirow{2}{*}{20\%}& Without ASI   & 15.82       \\
              & SAAP  & 16.35       \\
              \hline
\multirow{2}{*}{50\%}& Without ASI  & 31.73       \\
              & SAAP  & 32.41       \\
              \hline
    \end{tabular}
\end{table}
 
\begin{table}
\centering
  \renewcommand{\arraystretch}{1.25}
  \caption{Ablation Study for Efficient Group-Wise Fine-Tuning}
  \label{tab: ablation_finetune}
  
  \begin{tabular}{*{5}{c}}
  \hline
      Pruning & \multirow{2}{*}{Method} & Memory & \multirow{2}{*}{Params} & \multirow{2}{*}{Tokens/s}      \\
      Ratio & & (MiB)\\
    \hline
    0\%           & LLaMA-7B    & 12884.5 & 6.74B  & 25.84         \\
    \hline
    \multirow{3}{*}{20\% }         & SAAP (LoRA)  & 10286.5 & 5.38B  & 33.57(↑30\%) \\
                  & SAAP (QLoRA) & 10207.4 & 5.34B  & 34.51(↑34\%) \\
                  & SAAP        & 10055.7 & 5.26B  & 37.15(\textbf{↑44\%}) \\
                  \hline
    \multirow{3}{*}{50\%}          & SAAP (LoRA)  & 6285.9  & 3.29B  & 41.95(↑62\%) \\
                  & SAAP (QLoRA) & 6160.3  & 3.22B  & 42.55(↑65\%) \\
                  & SAAP        & 5940.8  & 3.12B  & 45.72(\textbf{↑77\%})\\
                  \hline
    \end{tabular}
\end{table}

\subsection{Ablation Study}
We perform an ablation study on SAAP’s three main components: the adaptive importance assessment, the adaptive structure search, and efficient group-wise fine-tuning. Additionally, we evaluate the impact of varying the number of calibration samples.

{\bf 1) Adaptive importance assessment.} The design of the importance estimation metric is crucial in determining which weights of an LLM are redundant and can be pruned without significantly degrading performance. The adaptive structure search part of our SAAP includes an innovative module, adaptive stability indicator, which integrates the comprehensive evaluation method of block importance judgment and volatility. We validate this approach through three distinct experimental methods.
\begin{itemize}
    \item Separate Cal: Use the original coarse-grained and fine-grained importance estimation methods, do not fuse their results, and calculate their relative fluctuations separately. By doing so, it serves as a baseline, allowing us to assess the impact of integrating these metrics. 
    \item Weighted Fusion: Simply weigh the importance of coarse-grained and fine-grained weights, and calculate the relative volatility of the weighted results. Instead of using the proposed adaptive importance fusion method, the results are directly calculated. 
    \item SAAP: The method proposed in this paper uses adaptive importance fusion metric.
\end{itemize}

In our experiment, we use the LLaMA-7B model with pruning ratios of 20\% and 50\%, respectively, and use the perplexity indicator on the WikiText2 dataset for evaluation. The final results are shown in Table \ref{tab: ablation_importance_estimation}. From Table \ref{tab: ablation_importance_estimation}, we can see that our proposed method has better results, and it can be found that directly fusing and weighting the importance of coarse-grained and fine-grained weights has a counter-effect and reduces the model's performance after pruning.

{\bf 2) Adaptive structure search.} To unify the differences in importance scores of each layer and module and reduce the impact of layered pruning on model performance, we propose adaptive stability indicator (ASI). To evaluate its effectiveness, we use LLaMA-7B for experiments, using pruning ratios of 20\% and 50\% respectively, and evaluate the perplexity metric on the WikiText2 dataset. The results are shown in Table \ref{tab: ablation_ASI}. It can be seen from the table that the proposed volatility-based metric as a pruning criterion can greatly improve the accuracy of the model.
\begin{figure}
    \centering
    \includegraphics[width=0.9\linewidth]{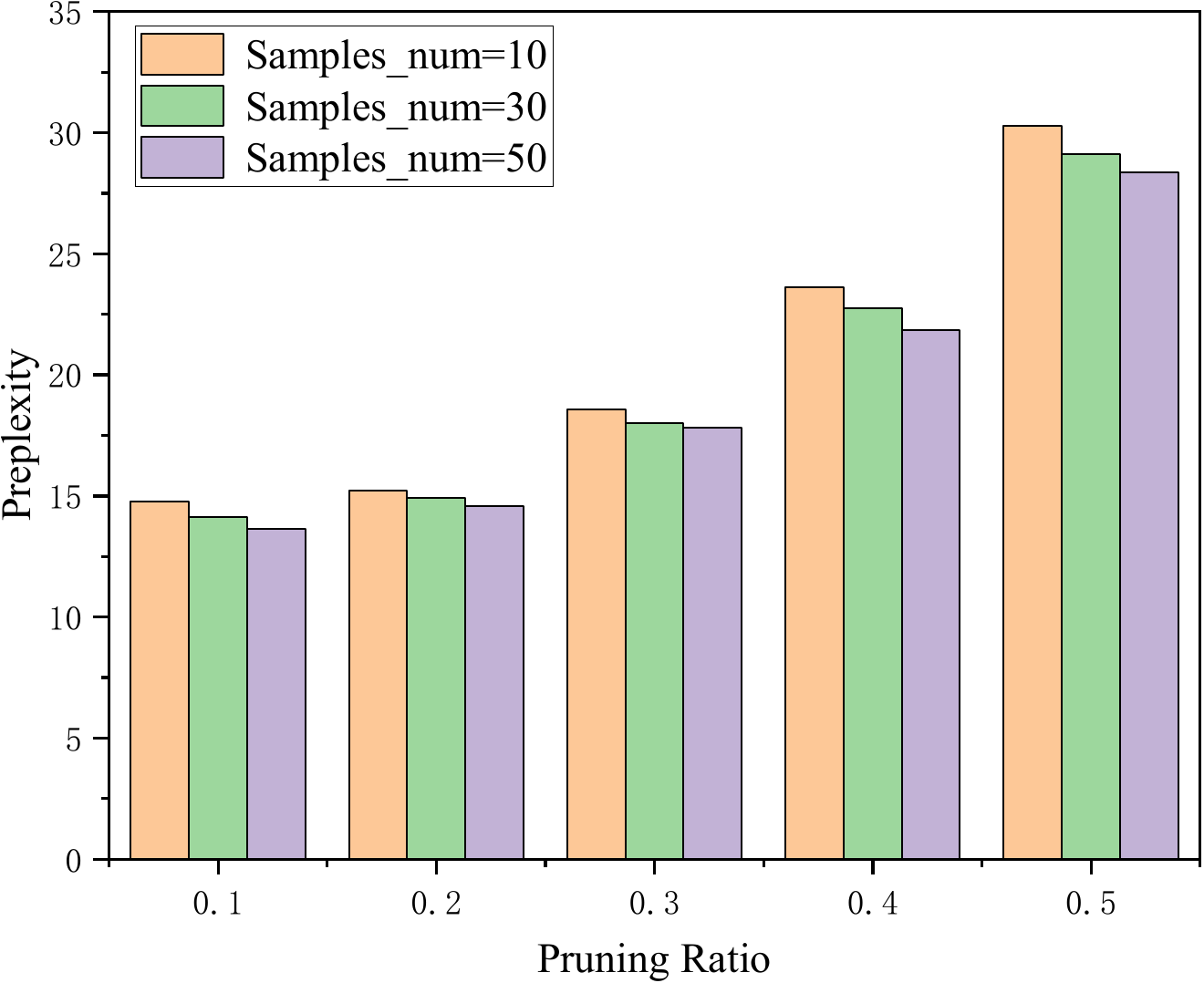}
    \caption{Ablation study for calibration sample numbers.}
    \label{fig different_ratio}
\end{figure}

{\bf 3) Efficient group-wise fine-tuning.} To verify the efficient group-wise fine-tuning, we replace this part with LoRA \cite{ref38} and QLoRA \cite{ref39} for testing. Table \ref{tab: ablation_finetune} shows the statistics of the 7B model in our experiment, including parameter count, memory requirements, and tokens per second. We use a single RTX3090 to perform the above test on the wikitext2 test set. From Table \ref{tab: ablation_finetune}, we can see that the proposed efficient group-wise fine-tuning can significantly improve the inference speed of the model.

{\bf 4) Numbers of calibration samples.} We use 10, 30, and 50 calibration samples for experiments, where the calibration samples are randomly selected from Bookcorpus \cite{ref56}. The experimental results are shown in Fig. \ref{fig different_ratio}. By adding some random calibration samples, the performance of the pruned model can also be improved.

\subsection{Discussion}
The SAAP method has demonstrated significant advantages in pruning LLMs. Through extensive testing on multiple LLMs, experimental results show that SAAP successfully reduces the number of model parameters, increases inference speed, and decreases memory usage, all while maintaining model performance. 

Firstly, the experimental results indicate that SAAP can maintain high model accuracy across different pruning ratios, especially at higher pruning ratios (50\%), where SAAP exhibits superior performance compared to other baseline methods. This advantage is primarily attributed to the adaptive importance fusion metric and adaptive structure search strategies employed by SAAP, which more precisely identify and remove redundant structures while retaining critical components essential for model performance. However, it was also observed that as the pruning ratio increases further, SAAP, though still leading, experiences significant absolute performance loss. 

Secondly, while SAAP excels in reducing model complexity and enhancing inference efficiency, its performance on certain datasets is slightly lower than that of existing methods. This phenomenon may be related to the number of random samples used in the experiments. The sample size of random sampling may not fully represent the characteristics and complexity of the entire dataset, potentially leading to SAAP's failure to capture the dataset's diversity in some cases. 

Moreover, SAAP's success heavily relies on its efficient group-wise fine-tuning strategy, which not only boosts the model’s inference speed but also achieves model quantization and low-rank decomposition without significantly compromising accuracy. However, this strategy might yield varying results across different model architectures, particularly in models with larger parameter scales. 

\section{Conclusion}
In this paper, we presented an efficient LLM pruning method called SAAP. It incorporated an adaptive importance metric in the estimation stage and used the importance fluctuation index as the evaluation criterion for adaptive structure search, thereby achieving effective pruning performance. In the recover stage, we developed a group-wise fine-tuning strategy to combine low rank and quantization efficiently. Through extensive experiments, we demonstrated the effectiveness of the proposed SAAP method, which achieved better inference quality and faster inference speed than several state-of-the-art baseline methods. Our work offered a novel perspective for LLM pruning, promising to achieve efficient and scalable LLM deployment in future intelligent applications.

\vfill
\end{document}